\documentclass[runningheads]{llncs}
\usepackage{graphicx}
\usepackage{tabularx} 
\usepackage{amsmath}
\usepackage{amssymb}
\usepackage[linesnumbered]{algorithm2e}
\begin{document}
\title{Data-Driven Simulation of Ride-Hailing Services using Imitation and Reinforcement Learning}
%
%
\newcommand*\samethanks[1][\value{footnote}]{\footnotemark[#1]}
\author{Haritha Jayasinghe\inst{1}\thanks{The first three authors have equal contribution} \and
Tarindu Jayatilaka\inst{1}\samethanks \and
Ravin Gunawardena\inst{1}\samethanks \and
Uthayasanker Thayasivam\inst{1}}
\authorrunning{Jayasinghe, Jayatilaka, Gunawardena, and Thayasivam}
%
\institute{University of Moratuwa, Moratuwa 10400, Sri Lanka\\
\email{\{haritha.16, tarindu.16, ravinsg.16, rtuthaya\}@cse.mrt.ac.lk }\\
\url{http://www.cse.mrt.ac.lk/}}
\maketitle              

\begin{abstract}
The rapid growth of ride-hailing platforms has created a highly competitive market where businesses struggle to make profits, demanding the need for better operational strategies. However, real-world experiments are risky and expensive for these platforms as they deal with millions of users daily. Thus, a need arises for a simulated environment where they can predict users' reactions to changes in the platform-specific parameters such as trip fares and incentives. Building such a simulation is challenging, as these platforms exist within dynamic environments where thousands of users regularly interact with one another. This paper presents a framework to mimic and predict user, specifically driver, behaviors in ride-hailing services. We use a data-driven hybrid reinforcement learning and imitation learning approach for this. First, the agent utilizes behavioral cloning to mimic driver behavior using a real-world data-set. Next, reinforcement learning is applied on top of the pre-trained agents in a simulated environment, to allow them to adapt to changes in the platform. Our framework provides an ideal playground for ride-hailing platforms to experiment with platform-specific parameters to predict drivers’ behavioral patterns. 

\end{abstract}

\section{Introduction}

Within less than a decade, internet-based ride-hailing platforms have managed to penetrate all global markets. However, all of these platforms are struggling with low or negative profit margins, implying that these platforms must be able to fine-tune parameters within their control to maximize profits. These parameters include surge pricing \cite{journals/corr/abs-1905-07544}, commissions, service fees, driver incentives, discounts, information available to the driver before accepting a trip (trip distance, payment method), and so on.

Experimenting with such parameters in a live platform is unrealistic due to; the high cost of running such trials, the in-adaptability of findings from one region to another, the exceedingly large number of trials required to search through combinations of multiple parameters, and the negative reactions from riders or drivers towards the changes and inconsistencies in the platform.

However, as these platforms begin to scale and offer more complex features, simulating them is essential to predict the behavior of agents---drivers and riders--- under different pricing and incentive strategies. Thus, the necessity arises for a simulated platform where such experiments can be conducted. The simulation should precisely predict how the agents react to different combinations of the above-mentioned parameters.

Unfortunately, realistically simulating the driver and passenger behavior within the platform is challenging as these platforms exist within highly volatile environments susceptible to change from external factors such as weather, traffic, or fuel prices. Existing simulations fail to capture precise driver behavior as they either utilize analytical models \cite{bailey1987} or attempt to simplify the objective function of drivers into simple reward functions \cite{gao2018optimize,lin2018}. Furthermore, any simulation requires the selection of certain parameters for precise modeling at the cost of abstracting away remaining parameters. The selection of the ideal parameter set for modeling within the simulation is a non-trivial task as it directly affects the accuracy of the predictions offered by the system.

This paper presents a novel approach to realistically mimic and predict the trip acceptance behavior of drivers on ride-hailing platforms. We propose a data-driven imitation learning approach to mimic driver behaviors which overcomes the aforementioned challenges of modeling driver behavior. Our model is portable across platforms as it can be adapted by merely changing the data set. Furthermore, deep reinforcement learning allows us to predict how agents change their behavior when the platform-specific parameters are changed using the reward function. Our prescriptive framework predicts the probability distribution of driver behavior on a macro scale in response to such changes.

We make the following contributions in this paper.
\begin{itemize}
\item We present a data-backed framework to simulate ride-hailing platforms and their users.
\item We demonstrate how imitation learning can be utilized to mimic drivers’ behaviors in a ride-hailing platform. 
\item We show how reinforcement learning agents pre-trained with behavioral cloning can be used to predict changes in driver behaviors in response to changes in the platform.
\end{itemize}

\section{Related Work}
The earliest efforts in modeling taxi-cab services were based on concepts such as linear programming and statistical modeling.  To investigate the relationship between the demand and the number of cabs to dispatch while overcoming the limitations in the previous methodologies, Bailey and Clark \cite{bailey1987} introduced the concept of simulated environments in ride modeling. The same researchers developed an event-based simulation \cite{bailey1992} to simulate taxi dispatching to reinforce the idea behind fleet size and performance. Despite the efforts, due to the lack of computational power during the era, the developed models were confined to use a minimal amount of data and resources, limiting their performance.

Machine learning is a more modern approach to model taxi cab services. Rossi et al. \cite{rossi2019modelling} used RNNs to predict the next location of taxi drivers. Lin et al. \cite{lin2018} worked on capturing the complicated stochastic demand-supply of taxi service variations in high-dimensional space using reinforcement learning. Gao et al. \cite{gao2018optimize} also used reinforcement learning to optimize taxi driving strategies to discover new operations decisions. Shou et al. \cite{SHOU202091} used an imitation learning approach to optimize passenger seeking behavior in ride-hailing platforms. However, all of the above primarily focus on mobility or demand and fleet size management compared to the drivers’ trip acceptance decisions, which is the primary focus of our work.

In terms of reinforcement learning, there are value-based and policy-based reinforcement learning algorithms. A prime example of a value-based algorithm is deep Q learning \cite{Mnih2015}, a model-free reinforcement learning approach commonly used in practical applications. There are also additional variations on DQ networks such as Categorical DQN \cite{bellemare2017}, which increases training stability. In contrast to value-based approaches that attempt to maximize cumulative reward, a policy-based approach such as the REINFORCE algorithm \cite{sutton1999} attempts to learn the best policy or set of actions to follow. In addition, there are also policy + action-based approaches such as actor-critic networks \cite{konda2000} and A2C networks \cite{mnih2016} which combines both of these strategies.

A major challenge of reinforcement learning is reward engineering. To solve this, we turn to imitation learning, particularly behavioral cloning, where the agent learns from real observations rather than through a reward function. Torabi et al. \cite{IJCAI2018-torabi} use observations first to learn a model and then use the learned model to infer the missing actions to improve the imitation policy. Goecks et al. \cite{goecks2020integrating} take this further by following behavioral cloning with reinforcement learning to take advantage of both approaches. It allows to capture the behavior of the agent and then improve the policy without additional observations. We adopt a similar approach in this paper.

\section{Methodology}

We propose a data-driven framework consisting of driver agents and a reinforcement learning environment. The driver agents are initially trained through imitation learning. Next, they are trained to adapt to varying platform parameters through a reinforcement learning approach, through interactions with the reinforcement learning environment. 

\begin{figure}
\centering
\includegraphics[width=\textwidth]{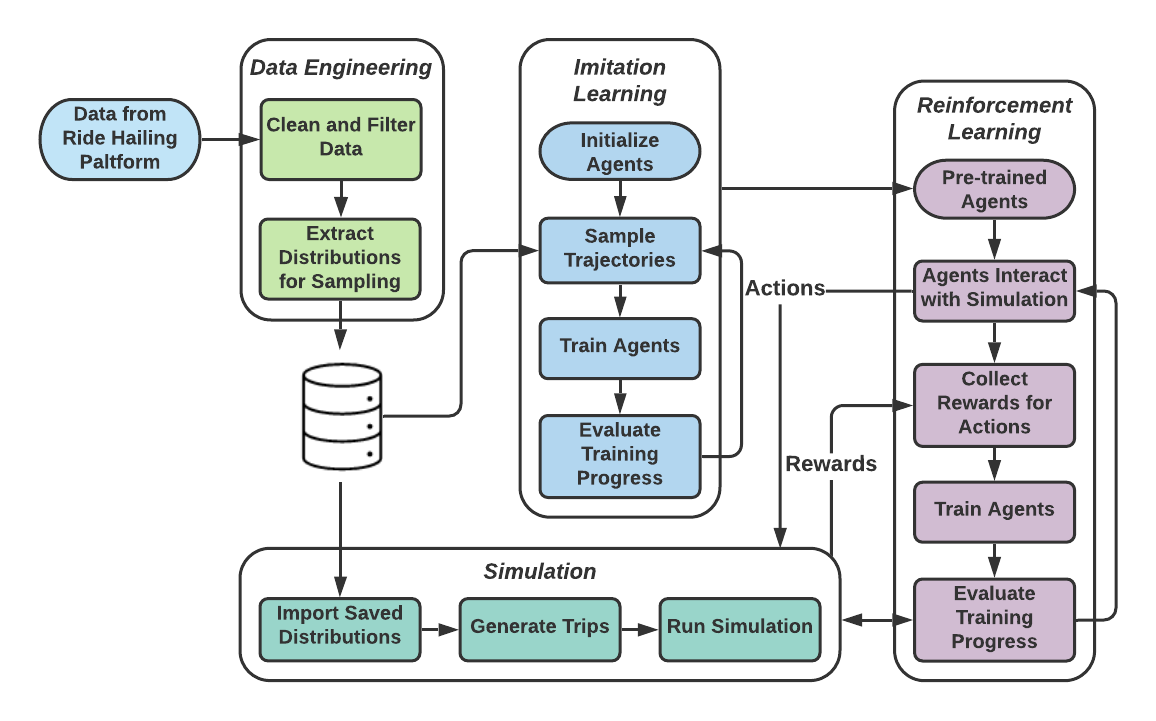} 
\caption{System Architecture}
\label{figsysarch}
\end{figure}

\subsection{Data}

For our work, we use a data set provided to us by a privately owned ride-hailing platform based in Sri Lanka. The data set consists of all the rides generated from 1\textsuperscript{st} February 2020 - 29\textsuperscript{th} February 2020 in the Colombo Metropolitan Area---a sprawling region with a population of approximately 6 million people occupying 1,422 square miles. Within the period, nearly 6.7 million trip requests have been prompted to the 28,982 drivers on the platform. Each trip is represented as a vector with the fields: driver id, trip id, trip created time, driver assigned time, trip accept/reject time, pickup time, pickup latitude, pickup longitude, drop latitude, drop longitude, distance to pick up, trip distance, status, and payment method. We only consider trips that occur in the busiest urban areas of Colombo. We also eliminate data that were duplicated, or erroneous, including trips with missing values. Around 5.6 million trips remained after cleaning, selection, and pre-processing of the data set.

For our simulation to accurately reflect the real world, the above data set is incorporated into the system using sampling techniques. We treat the coordinates of each location as two random variables \(L_x \text{ and } L_y\) distributed across our region. Each random variable is described using its own cumulative distribution function: \(F_{L_x} \text{ and } F_{L_y}\). We assume that the number of data points we have is a sufficient approximation for the actual CDF of the distribution. Finally, we use the \textit{inverse transformation method} to generate values for \(L_x\) and \(L_y\). We can sample values for \(L\) using \(F_L^{-1}\), where \(F_L\) is a common representation of \(F_{L_x} \text{ and } F_{L_y}\). Similarly, we create another random variable \(D\) to sample trip distances.

\subsection{Simulation Framework}

The simulation framework is composed of driver agents and the reinforcement learning environment.

\subsubsection{Reinforcement Learning Environment}

The RL environment comprises the interactions between drivers, riders, and the ride-hailing platform. The ability of the simulation to reflect reality directly affects the performance of the agents. Hence, the primary goal of building this environment is to incorporate real-world information and replicate the operations taking place within the ride-hailing platform.

From here on, we abstract the behavior of riders and refer to them as rides for modeling purposes. A ride is a vector, \(R = (r_x^p, r_y^p ,r_x^d, r_y^d, r_d)\), of variables sampled from the distributions we extract from the data set. Each ride has pickup and drop coordinates (\([r_x^p,r_y^p ], [r_x^d, r_y^d]\)) and a ride distance (\(r_d\)). All the geographic coordinates are in a rectangular grid of size \(M_h \times M_w\).

We only simulate a fraction of the total available rides to increase the throughput of the system and to optimize the training time of the agents. We use a scaling factor of 35 to scale down the number of rides generated per week to around 20,000. To maintain the distribution coherent when the scaled ride count per minute is below 1, which accounts for 25-30\% of the total time, a probabilistic approach is taken instead of naively rounding off the decimal places, for better accuracy. The scaled ride count is rounded up with a probability equivalent to its fractional part. Consequently, the expected value of the number of trips per minute becomes equal to the scaled ride count. 

\subsubsection{Ride Generation Algorithm}

A ride \((R)\) is described using the pickup/ drop coordinates and the ride distance. The time of day \((t_d)\) decides the frequency of the rides being generated. For each minute, the number of trips per minute \((N_t)\) is sampled from the time distribution \((T)\) using the time of day \((t_d)\). Using the number of trips per minute \((N_t)\), location distribution \((L)\), and distance distribution \((D)\), Algorithm~\ref{algo:GenerateRides} generates trips for the simulation.

\begin{algorithm}
\DontPrintSemicolon 
\KwIn{
    Map Size \([M_h, M_w]\) \\
    Random Variables \(L_x, L_y, D\) \\
    Number of Rides Per Minute \(N_t\)
}
\KwOut{
    Queue of Ride Objects  $\langle r_1, r_2, \ldots, r_n \rangle$
}
$rides \gets [\quad] $\;
$ride\_no \gets 0 $\;
\While{$ride\_no < N_t$} {
    $(r_x^p , r_y^p) \gets (L_x(\omega), L_y(\omega))$ for $\omega \in \Omega$\;
    $(r_x^p , r_y^p) \gets (r_x^p , r_y^p) + (\mathcal{U}(-\epsilon,\epsilon), \mathcal{U}(-\epsilon,\epsilon))$\;
    $r_x^p \gets max(0, min(M_w, r_x^p))$\;
    $r_y^p \gets max(0, min(M_h, r_y^p))$\;
    $r_d \gets D(\omega)$ for $\omega \in \Omega$\;
    $(r_x^d , r_y^d) \gets $ calculate drop location \;
    \While{not $0 < r_x^d < M_w$ and $0 < r_y^d< M_h$} {
        $r_d \gets \frac{r_d}{2}$\;    
        $(r_x^d , r_y^d) \gets $ calculate drop location \;
    }
    $r_i \gets (r_x^p , r_y^p, r_x^d , r_y^d, r_d)$\;
    $rides.insert(r_i)$\;
    $ride\_no \gets ride\_no + 1$\;
}
\Return{$rides$}\;
\caption{{\sc GenerateRides} Generates Rides for the Simulation }
\label{algo:GenerateRides}
\end{algorithm}

The vector \([r_x^p, r_y^p]\) which is sampled from \(L\) (\textit{line 4}), represents the latitude and longitude (\(x, y\)) coordinates of the pickup location. We add a small amount of random noise uniformly distributed between \((-\epsilon, \epsilon)\) to make the location different every time (\textit{line 5}). The coordinates have an upper bound of \(M_h\) or \(M_w\), and a lower bound of 0. The trip distance \(t_d\) is sampled from \(D\) (\textit{line 8}). Now, we can draw a circle of radius \(t_d\) having the center as \([r_x^p, r_y^p]\). A random point on the circumference of this circle is selected as the drop location, \([r_x^d, r_y^d]\). If all possible locations are outside the grid, \(t_d\) is halved and the process is restarted (\textit{lines 10-12}). Once the drop location is identified, the ride is initialized and pushed to a queue to be processed by the simulation.

\subsubsection{Agent Modelling Approach}

We model the ride accept/reject decision for each driver in the simulation using an agent trained using a combination of imitation learning and deep reinforcement learning. Specifically, we use a two-step approach. We first train the agent to clone the behavior demonstrated by the drivers in our data-set, allowing the agent to learn the inherent behavioral patterns of the drivers. Next, we add a layer of reinforcement learning, utilizing rides generated from our simulation, thereby adapting the agent to the platform-specific parameters we define within the simulation. This approach is described in detail below.

We initially utilize the data-set to generate demonstrations, which are treated as independent and identically distributed (i.i.d.) samples, to follow a behavioral cloning approach. We utilize a deep Q network \cite{Hausknecht2015DeepRQ} for training purposes, and provide the demonstrations gathered above as trajectories for the training process. 

A single training trajectory \([s,a,s',r]\) consists of; \(1.\) environment state immediately before the action \((s)\), \(2.\) action \((a)\), \(3.\) environment state immediately after the action \((s')\), \(4.\)reward associated with the action \((r)\).

We expose the environment state to our agent via the following observations; \(1.\) the distance to pick-up location from current driver location, \(2.\) the trip distance, \(3.\) the time of day (useful to infer information regarding peak / off-peak time and demand patterns), \(4.\) the number of trips to complete till weekly reward is achieved, \(5.\) trip destination, \(6.\) the idle time since last completed trip (useful for inferring the opportunity cost of remaining idle).

We posit that by providing observations such as \emph{Trips Left till Weekly Reward}, we can overcome limitations imposed by the i.i.d. assumption of behavioral cloning approaches in simulating Markov Decision Processes (MDP), as such observations can provide an agent a rudimentary understanding of its state history. 

Secondly, we move to a reinforcement learning approach to train the agent on the simulation with a set of platform parameters. These parameters include various reward factors, which can be adjusted as required. During this step, the agent interacts directly with the simulation. We utilize a relatively lower exploration rate and early stopping to ensure that the agent retains the patterns learned during the imitation learning phase. 

The basic process of the reinforcement learning system is as follows.

\begin{itemize}
\item Whenever a trip is generated and offered to a driver(agent), the agent is provided with a set of observations that describe the environment.
\item Based on the observations the agent is required to make a decision. (accept/reject trip)
\item The agent is given a reward for the action taken.
\end{itemize}

Drivers take actions to maximize their rewards according to the concepts of Game Theory. Thus, a key factor required for realistic simulation of driver behavior is the ability to accurately represent the reward function of the driver. Ideally, the reward function should be capable of representing the fact that human agents are not perfectly rational, however using traditional reinforcement learning, we are restricted to representations of rational behavior only. The reward used for training the agents is calculated as a function of trip distance(td), pickup distance(pd), opportunity cost (oc), and weekly reward(wr).

\begin{align*}
&Q(s, a) := Q(s,a) + \alpha( r + \gamma max_{a’}Q(s’, a’)  - Q(s, a))
\stepcounter{equation}\tag{\theequation}\label{eq1} \\
&r = w*(fare/km * td) - x*(cost/km * (td + pd) - y*oc + z*wr \stepcounter{equation}\tag{\theequation}\label{eq2} \\
\end{align*}

The opportunity cost is a penalty calculated based on the driver’s active time, which represents the value of the time invested into the platform by the driver, and the weekly reward is an additional reward present in the platform to motivate drivers. Each driver is given a weekly goal based on his past performance, which he must meet to gain this reward. \(w\), \(x\), \(y\), and \(z\) represent weights that scale an agent's preference for different types of rewards. The agent attempts to select actions that maximize the overall expected value of the decisions taken. This is achieved by iteratively updating the expected value (\(Q\) value) for each training example, as given in the below equation. \(\gamma\) represents a discounting factor for future rewards, and \(s\) represents the state (sequence of actions and observations), while \(a'\) and \(s'\) denote all actions and the resultant state of all actions respectively. \cite{Hausknecht2015DeepRQ}

While the most obvious method to model individual drivers would be to use separate DQN agents to represent each driver, this is impractical due to the large number of training examples required for each agent. For simplicity, we train a single agent for all drivers and utilize an experiment setup with 50 drivers.

For improved stability, we utilize a categorical deep Q network \cite{bellemare2017} instead of a vanilla DQN, where a Q value distribution is learned as opposed to a single Q value, which results in a more stable learning process due to reduced chattering, state aliasing, etc.

\section{Experimental Setup}

Our experimental setup consists of two main sections. The simulation environment is evaluated based on the accuracy of the generated rides in their ability to reflect real-world data. The agent learning method is evaluated by both the agent's ability to mimic driver behavior, as well as the agent's ability to adapt to platform parameters. This is achieved by comparing the behavioral changes of the drivers in response to changes in parameters. Our results reflect the success of the system in predicting the agent behavior at a macroscopic level.

In our setup, we train a single imitation learning agent using around 4.2 million trips, for 150 iterations with a replay buffer containing 1000 trajectories. A single trajectory represents all trip acceptance/rejection actions of a given driver over three weeks, as the final week of the data set was excluded for evaluation purposes.

During the reinforcement learning setup, we train for a further 50 iterations, where a single iteration consists of all trajectories collected in a single run of our simulation. A single run involves a 1 week run time. We evaluate the adaptive behavior of the agent through the following experiment setup. The agent is first trained through the imitation learning process. Subsequently, copies of the agent are trained on variations of the simulation with different platform parameters. Post-training, we observe the behavior of the agents to understand behavioral changes induced by the parameter changes during the reinforcement learning phase.

\section{Results}

For the purpose of evaluating the efficacy of our simulation, Table~\ref{tab1} compares the predicted number of trips for the final week by the simulation, against the actual number of trips from the data-set.  

\begin{table}[h]
\caption{Comparison of predicted trip count against the actual trip count. The upper and lower bounds are the intervals of 95\% confidence level.} 
\label{tab1}
\centering
\begin{tabularx}{\columnwidth}{|X|r|r|r|r|r|}
        \hline
        {\centering Day}     & {\centering Predict. (Mean)}  & {\centering Lower Bound}  & {\centering Upper Bound}  & {\centering Actual} & {\centering {$\delta(\%)$}}    \\
        \hline
        Mon     & 110,852    & 110,162    & 111,542    & 100,905  & +9.858 \\
        Tue     & 106,807    & 106,141    & 107,473    & 108,280   & -1.360 \\
        Wed     & 115,947    & 115,487    & 116,407    & 122,377   & -5.254 \\
        Thu     & 116,370    & 115,589    & 117,151    & 127,664   & -8.847 \\
        Fri     & 149,515    & 148,561    & 150,469    & 145,924   & +2.461 \\
        Sat     & 83,263     & 82,746     & 83,780     & 88,365    & -5.774 \\
        Sun     & 57,442     & 57,054     & 57,830     & 58,734    & -2.200 \\
        \hline
\end{tabularx}
\end{table}

\begin{figure}[h!]
\centering
\includegraphics[width=0.97\columnwidth]{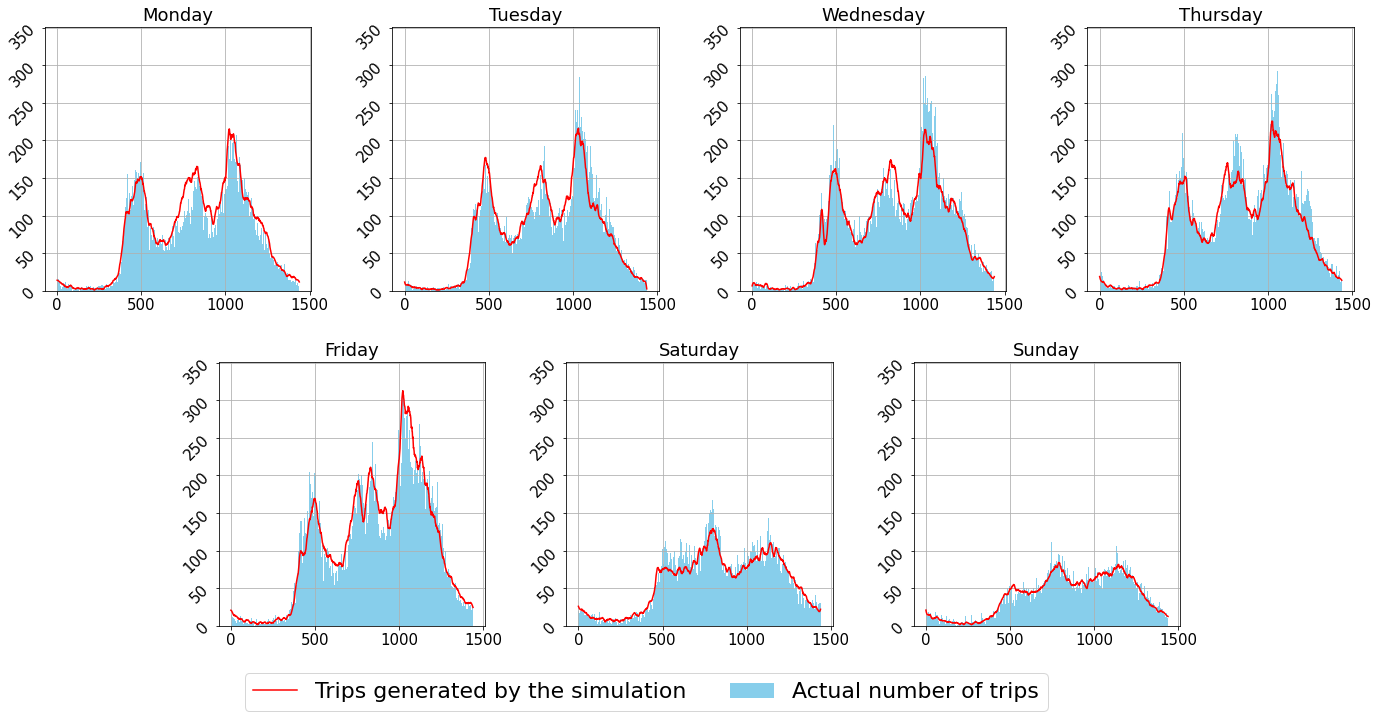} 
\caption{Distribution of the predicted number of trips generated by the simulation (red) against the actual number of trips (blue) for each day of the week in the last week of February}
\label{fig1}
\end{figure}

We observe that the simulation is successful in predicting the demand of the economy to a significant extent. However, it is not perfect. This could be due to other factors we did not take into consideration: weather, traffic, special events, holidays, etc. The notable gap between the predicted and actual ride count on Thursday reflects this notion. Since the following Friday was a public holiday in Sri Lanka, the demand for rides increased on Thursday evening.

Figure~\ref{fig1} shows the distribution of the actual demand for rides against the predicted demand for each day of the week. It can be seen that the simulation is successful in identifying the underlying demand patterns of the ride-hailing platform sufficiently. 

\begin{figure}[h!]
\centering
\includegraphics[width=0.97\columnwidth]{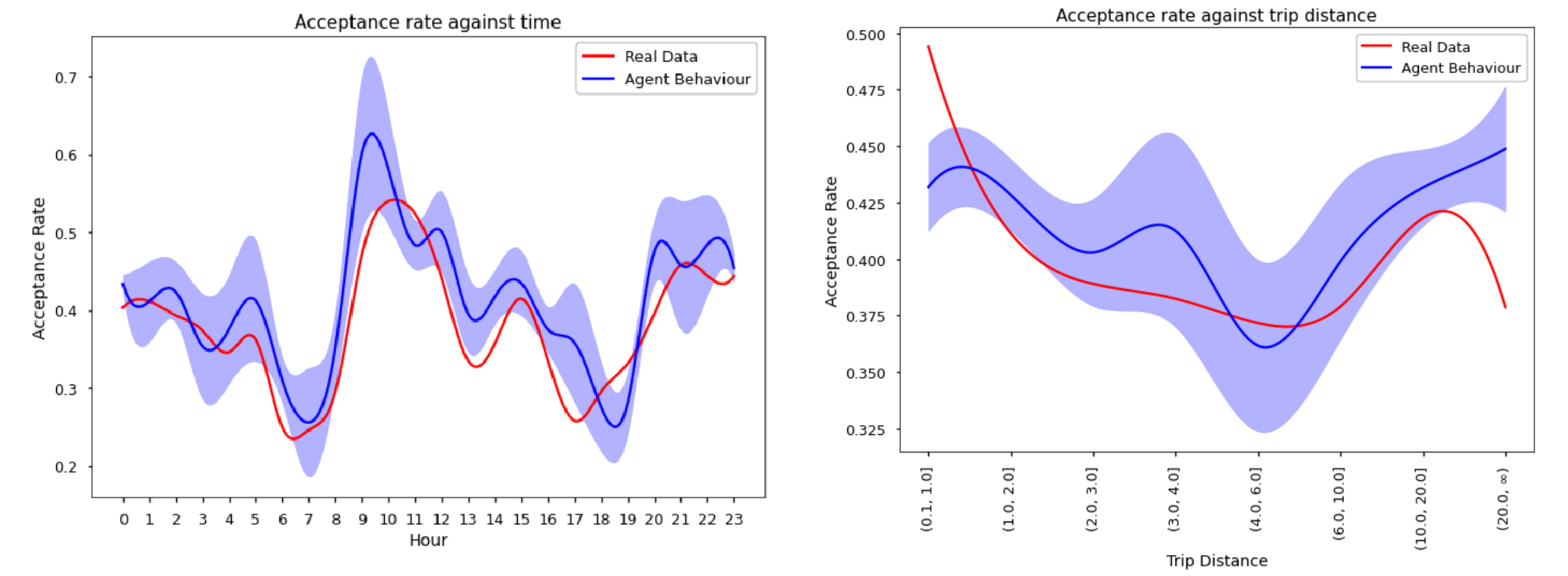} \caption{Comparison of the actual driver behavior and the agent behavior after training using Imitation Learning}
\label{IL}
\end{figure}

To compare the behavior of the drivers trained through imitation learning against the behavior of drivers from the real-world dataset, we compute the correlation between their acceptance rates. Figure~\ref{IL} shows that our imitation agents' behavior strongly correlates with the actual driver behavior with a Pearson correlation coefficient of 0.875.  

\begin{figure}[h!]
\centering
\includegraphics[width=0.97\columnwidth]{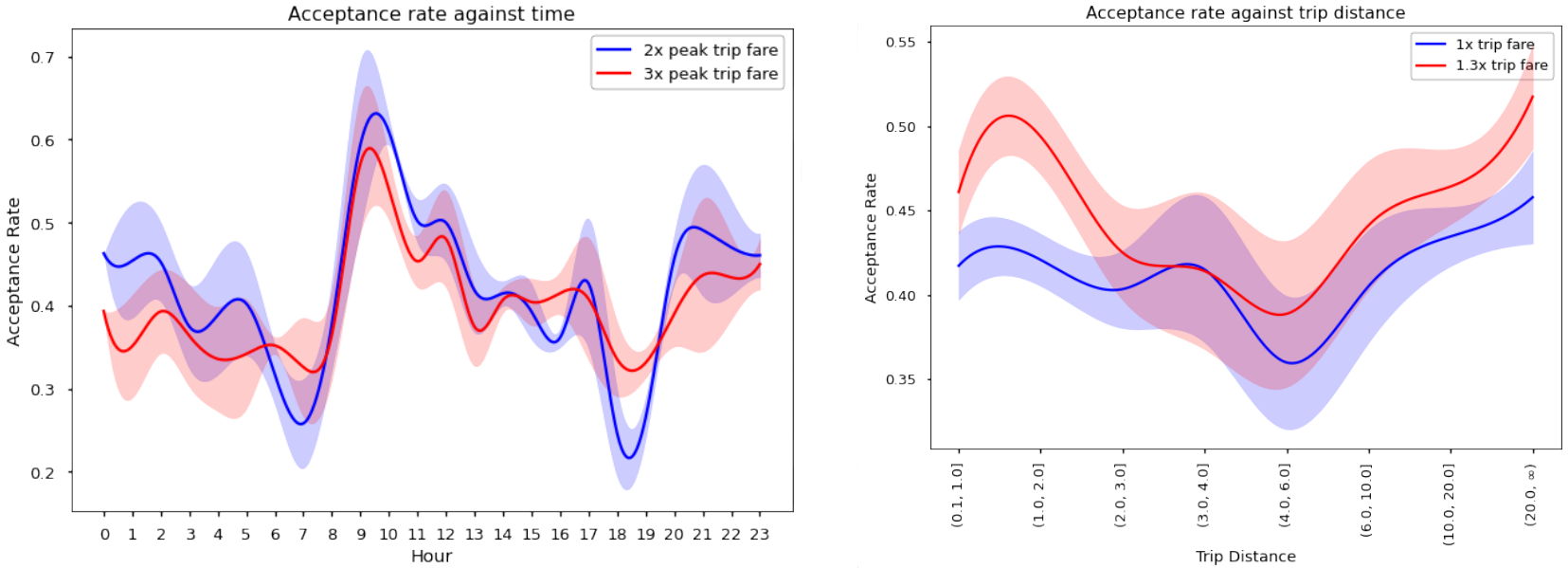} \caption{Comparison of agent behavior against time of day(hrs) with 2x and 3x peak trip fare(left), and Comparison of agent behavior against trip distance(km) with 1x and 1.3x trip fare(right)}
\label{figil1}
\end{figure}

When evaluating the performance of the reinforcement learning approach, Figure~\ref{figil1} (left) demonstrates the contrast between providing 2x vs 3x trip fare (per km, relative to off-peak rate) during peak hours, plotted against time of day. (6-8 hrs and 16-19hrs are defined as peak hours). As expected, the acceptance rate during peak hours has increased, however with the caveat of a slightly reduced acceptance rate during off-peak hours. Crucially, in practice, these parameters do not operate in isolation, and the gap between peak and off-peak rates can also be controlled through other parameters such as weekly rewards. Figure~\ref{figil1} (right) depicts the effect of increasing trip fare (per km) by 1.3x, where it can be observed that the acceptance rates increase noticeably, across all distances, as the trips become more profitable. 

\begin{figure}[h!]
\centering
\includegraphics[width=0.97\columnwidth]{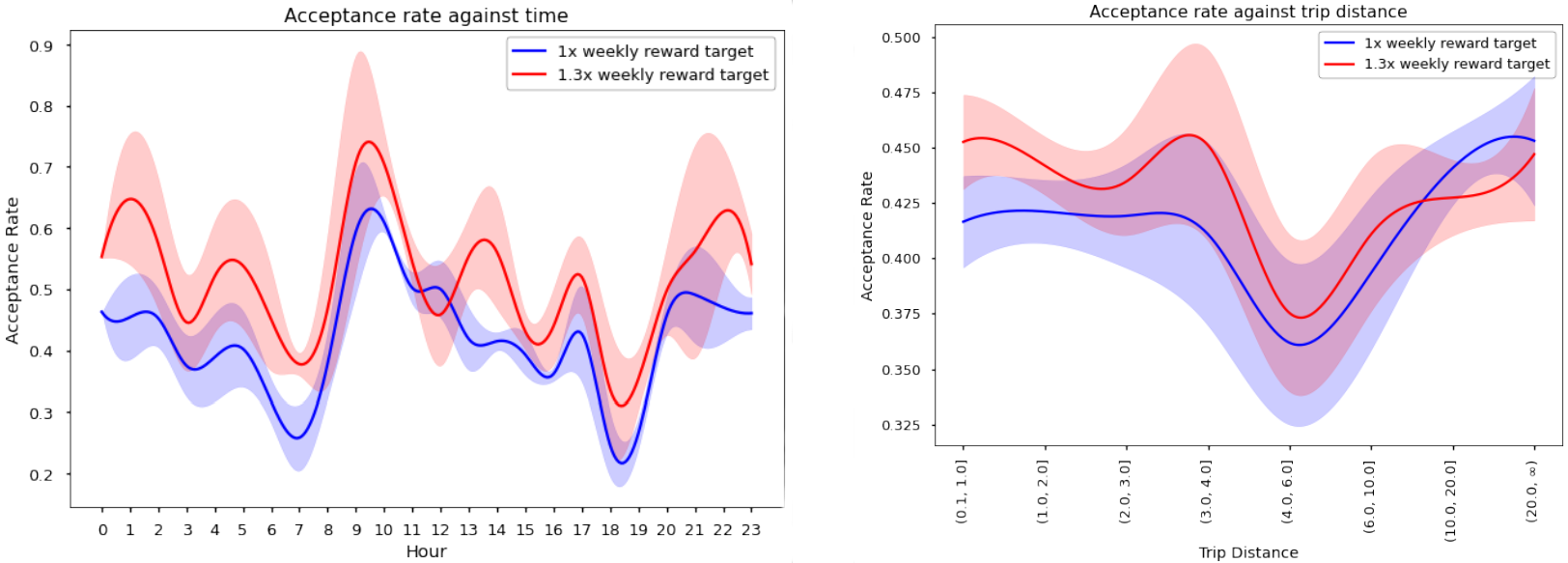} \caption{Comparison of agent behavior against time of day(hrs) (left) and agent behavior against trip distance(km) (right) with weekly 1x and 1.5x reward targets}
\label{figil2}
\end{figure}
 
Figure~\ref{figil2} demonstrates the contrast between setting the weekly reward target at  1x vs 1.3x the previous week’s performance. A notable increase can be observed in the general acceptance rates when reward targets are increased, particularly for trips with lower distances, as the profitability of these trips now increase. Interestingly, the higher reward target does not cause a notable increase in the lower acceptance rate during peak hours, implying that the drivers can cherry-pick trips and still maintain their goal. 

The above are two of the various platform parameters that can be tweaked. Since the behavioral response of drivers become more complicated as these parameters interact with one another, complicated behavioral patterns can be monitored by tweaking them in tandem, to identify parameters that maximize the KPIs of the platform.

\section {Conclusion}

We present a novel data-driven simulation framework that uses imitation learning and Categorical Deep Q-Networks to model driver behaviors in ride-hailing platforms. Our simulation successfully manages to predict the economy at a macroscopic level. We achieve a margin of error of less than 10\% in predicting the demand through the week and the total revenue generated by the platform. The quantitative results prove the ability of the DQN agent to imitate the behavior of drivers with a strong correlation coefficient of 0.875, against observations such as trip distance and time of day. Furthermore, we can observe how the agent behavior changes using reinforcement learning when we tweak the platform-specific parameters using in the simulated environment. Our framework demonstrates that any ride-hailing service, where assets and services are shared between a large number of individuals interacting with one another, can be simulated using imitation learning.

\section {Future Work} 

While our approach utilizes behavioral cloning to learn driver behavior, Inverse Reinforcement Learning presents a more advanced method to potentially capture a reward function that depicts driver behavior more effectively. Furthermore, we currently allow users to manually tweak a parameter and observe driver behavior. Zheng et al. \cite{zheng2020ai} utilize multi-agent reinforcement learning to simulate the behavior of participants as well as a rule-setting agent, demonstrating a method to create a system that can automatically determine the ideal rule set to maximize the utility of the system. This approach can be adapted to our use case, where an additional RL agent can be designed to tweak the parameters of the platform (incentives, information available to the driver at decision time, etc.) in between simulation episodes. Once trained, such a system could automatically identify the ideal set of parameters to maximize the revenue of the ride-hailing platform.

%
%
%
\bibliographystyle{splncs04}
\bibliography{bibfile.bib}

\end{document}